%% file: main.tex
\documentclass{article}

\PassOptionsToPackage{numbers}{natbib}
\usepackage[preprint]{nips_2018}

\usepackage[utf8]{inputenc} 
\usepackage[T1]{fontenc}    
\usepackage{hyperref}       
\usepackage{url}            
\usepackage{booktabs}       
\usepackage{amsfonts}       
\usepackage{nicefrac}       
\usepackage{microtype}      

\usepackage{algorithm}
\usepackage{algpseudocode}
\usepackage{subcaption}
\usepackage{enumitem}
\usepackage{amsmath}
\usepackage{amssymb}
\usepackage{stmaryrd}
\usepackage{dsfont}
\usepackage{multirow}
\usepackage{comment}
\usepackage{wrapfig}
\usepackage[font={small}]{caption}
\usepackage{amsthm}

\DeclareMathOperator*{\argmin}{argmin}
\DeclareMathOperator*{\argmax}{argmax}
\usepackage[colorinlistoftodos]{todonotes}

\newcommand{\cutsectionup}{\vspace*{-0.0in}}
\newcommand{\cutsectiondown}{\vspace*{-0.0in}}

\newcommand{\cutsubsectionup}{\vspace*{-0.0in}}
\newcommand{\cutsubsectiondown}{\vspace*{-0.0in}}

\newcommand{\cutparagraphup}{\vspace*{-0.0in}}

\newcommand{\gasil}{Generative Adversarial Self-Imitation Learning}
\newcommand{\GASIL}{GASIL}

\title{\gasil}

\author{
  Yijie Guo$^{*1}$ \texttt{  } \texttt{  }
  Junhyuk Oh$^{*1\dagger}$ \texttt{  } \texttt{  } 
  Satinder Singh$^1$ \texttt{  } \texttt{  }
  Honglak Lee$^{1,2}$ \\
  $^1$University of Michigan \\ $^2$Google Brain \\
  \texttt{\{junhyuk,guoyijie,baveja,honglak\}@umich.edu, \{junhyuk,honglak\}@google.com}
}

\begin{document}

\maketitle

\begin{abstract}
This paper explores a simple regularizer for reinforcement learning by proposing \textit{Generative Adversarial Self-Imitation Learning} (\GASIL{}), which encourages the agent to imitate past good trajectories via generative adversarial imitation learning framework. Instead of directly maximizing rewards, \GASIL{} focuses on reproducing past good trajectories, which can potentially make long-term credit assignment easier when rewards are sparse and delayed. \GASIL{} can be easily combined with any policy gradient objective by using \GASIL{} as a learned shaped reward function. 
Our experimental results show that \GASIL{} improves the performance of proximal policy optimization on 2D Point Mass and MuJoCo environments with delayed reward and stochastic dynamics.
\end{abstract}
\vskip -0.1in

\cutsectionup
\section{Introduction}
\cutsectiondown
Reinforcement learning (RL) is essentially a temporal credit assignment problem that amounts to figuring out which action in a state leads to a better outcome in the future. Different RL algorithms have different forms of objectives to solve this problem. For example, policy gradient approaches learn to directly optimize the RL objective (i.e., maximizing rewards), while value-based approaches (e.g., Q-Learning~\citep{watkins1992q}) indirectly optimize it by estimating long-term future rewards and inducing a policy from it. Policies optimized for different objectives many have different learning dynamics, which end up with different sub-optimal policies in complex environments, though all of these objectives are designed to maximize rewards. 

In this paper, we explore a simple regularizer for RL, called \textit{\gasil{}} (\GASIL{}). Instead of directly maximizing rewards, \GASIL{} aims to imitate past good trajectories that the agent has generated using generative adversarial imitation learning framework~\citep{ho2016generative}. \GASIL{} solves the temporal credit assignment problem by learning a discriminator which discriminates between the agent's current trajectories and good trajectories in the past, while the policy is trained to make it hard for the discriminator to distinguish between the two types of trajectories by imitating good trajectories. 
\GASIL{} can potentially make long-term temporal credit assignment easier when reward signal is delayed and sparse. 
\GASIL{} can be interpreted as an optimal reward learning algorithm~\citep{singh2009rewards,sorg2010reward}, where the discriminator acts as a learned reward function which provides dense rewards for the agent to reproduce relatively better trajectories. Thus, it can be used as a shaped reward function and combined with any RL algorithms.

Our empirical results on 2D Point Mass and OpenAI Gym MuJoCo tasks~\citep{brockman2016openai,Todorov2012MuJoCoAP} show that \GASIL{} improves the performance of proximal policy optimization (PPO)~\citep{Schulman2017ProximalPO}, especially when rewards are delayed. We also show that \GASIL{} is robust to stochastic dynamics to some extent in practice. 

\cutsectionup
\section{Related Work}
\cutsectiondown

\paragraph{Generative adversarial learning}
Generative adversarial networks (GANs)~\citep{goodfellow2014generative} have been increasingly popular for generative modeling. In GANs, a discriminator is trained to discriminate whether a given sample is drawn from data distribution or model distribution. A generator (i.e., model) is trained to ``fool'' the discriminator by generating samples that are close to the real data. This adversarial play allows the model distribution to match to the data distribution. This approach has been very successful for image generation and manipulation~\citep{radford2015unsupervised,reed2016generative,zhu2017unpaired}. 
Recently, \citet{ho2016generative} proposed generative adversarial imitation learning (GAIL) which extends this idea to imitation learning. In GAIL, a discriminator is trained to discrimiate between optimal trajectories (or expert trajectories) and policy trajectories, while the policy learns to fool the discriminator by imitating optimal trajectories. 
Our work further extends this idea to RL. Unlike GAN or GAIL setting, optimal trajectories are not available to the agent in RL. Instead, \GASIL{} treats ``relatively better trajectories'' that the policy has generated as optimal trajectories that the agent should imitate. 

\paragraph{Reward learning}
\citet{singh2009rewards} discussed a problem of learning an internal reward function that is useful across a distribution of environments in an evolutionary context. Simiarly, \citet{sorg2010reward} introduced \textit{optimal reward problem} under the motivation that the true reward function defined in the environment may not be optimal for learning, and there exists an optimal reward function that allows learning the desired behavior much quickly. This claim is consistent with the idea of reward shaping~\citep{ng1999policy} which helps learning without changing the optimal policy. There has been a few attempts to learn such an internal reward function without domain-specific knowledge in deep RL context~\citep{sorg2010reward,Guo2016DeepLF,zheng2018learning}. Our work is closely related to this line of work in that the policy does not explicitly optimize the reward given by the environment in \GASIL{}. Instead, \GASIL{} learns a discriminator which acts as an interal reward function that the policy should maximize. 


\paragraph{Self-imitation}
Learning a policy by focusing on past good experiences has been discussed. For example, episodic control~\citep{lengyel2008hippocampal,blundell2016model,pritzel2017neural} and the nearest neighbor policy~\citep{mansimov2017simple} construct a non-parametric policy directly from the past experience by retreiving similar states in the past and choosing the best action made in the past. Our work aims to learn a parametric policy from past good experiences. Self-imitation has been shown to be useful for program synthesis~\citep{liang2016neural,abolafia2018neural}, where the agent is trained to generate K-best programs generated by itself. Our work proposes a different objective based on generative adversarial learning framework and evaluates it on RL benchmarks. More recently, \citet{goyal2018recall} proposed to learn a generative model of preceding states of high-value states (i.e., top-$K\%$ trajectories) and update a policy to follow the generated trajectories. In contrast, our \GASIL{} directly learns to imitate past good trajectories without learning a generative model. \GASIL{} can be viewed as a generative adversarial extension of \textit{self-imitation learning}~\citep{oh2018self} which updates the policy and the value function towards past better trajectories. Contemporaneously with our work, \citet{gangwani2018learning} also proposed the same method as our \GASIL{}, which was independently developed. 
Most of the previous works listed above including ours require environments to be deterministic in order to guarantee policy improvement due to the bias towards positive outcome. Dealing with stochasticity with this type of approach would be an interesting future direction.

\cutsectionup
\section{Background}
\cutsectiondown
Throughout the paper, we consider a finite state space $\mathcal{S}$ and a finite action space $\mathcal{A}$. The goal of RL is to find a policy $\pi  \in \Pi :\mathcal{S} \times \mathcal{A} \rightarrow [0, 1]$ which maximizes the discounted sum of rewards: $\eta(\pi)=\mathbb{E}_\pi\left[\sum^{\infty}_{t=0} \gamma^t r_t\right]$ where $\gamma$ is a discount factor and $r_t$ is a reward at time-step $t$. 

Alternatively, we can re-write the RL objective $\eta(\pi)$ in terms of occupancy measure. Occupancy measure $\rho_\pi \in \mathcal{D} : \mathcal{S} \times \mathcal{A} \rightarrow \mathbb{R}$ is defined as $\rho_\pi(s,a) = \pi(a|s)\sum_{t=0}^{\infty}\gamma^tP(s_t = s|\pi)$. Intuitively, it is a joint distribution of states and actions visited by the agent following the policy $\pi$. It is shown that there is a one-to-one correspondence between the set of policies ($\Pi$) and the set of valid occupancy measures ($\mathcal{D}=\{ \rho_\pi : \pi \in \Pi\}$)~\citep{syed2008apprenticeship}. This allows us to write the RL objective in terms of occupancy measure as follows:
\begin{align}
\eta(\pi)=\mathbb{E}_\pi\left[\sum^{\infty}_{t=0} \gamma^t r_t\right]=\sum_{s,a}\rho_\pi(s,a)r(s,a).
\end{align}
where $r(s,a)$ is the reward for choosing action $a$ in state $s$.
Thus, policy optimization amounts to finding an optimal occupancy measure which maximizes rewards due to the one-to-one correspondence between them.

\cutsubsectionup
\subsection{Policy Gradient} \label{sec:pg}
\cutsubsectiondown
Policy gradient methods compute the gradient of the RL objective $\eta(\pi_\theta)=\mathbb{E}_{\pi_\theta}\left[\sum^{\infty}_{t=0} \gamma^t r_t\right]$. Since $\eta(\pi_\theta)$ is non-differentiable with respect to the parameter $\theta$ when the dynamics of the environment are unknown, policy gradient methods rely on the score function estimator to get an unbiased gradient estimator of $\eta(\pi_\theta)$. A typical form of policy gradient objective is given by:
\begin{align}
J_{\text{PG}}(\theta) = \mathbb{E}_{\pi_\theta} \left[\log \pi_\theta(a_t|s_t)\hat{A}_t \right]
\label{eq:pg}
\end{align}
where $\pi_\theta$ is a policy parameterized by $\theta$, and $\hat{A}_t$ is an advantage estimation at time $t$. Intuitively, the policy gradient objective either increases the probability of the action when the return is higher than expected ($\hat{A}_t > 0$) or decreases the probability when the return is lower than expected ($\hat{A}_t < 0$). 

\cutsubsectionup
\subsection{Generative Adversarial Imitation Learning} \label{sec:gail}
\cutsubsectiondown
Generative adversarial imitation learning (GAIL)~\citep{ho2016generative} is an imitation learning algorithm which aims to learn a policy that can imitate expert trajectories using the idea from generative adversarial network (GAN)~\citep{goodfellow2014generative}. More specifically, the objective of GAIL for maximum entropy IRL~\citep{ziebart2008maximum} is defined as:
\begin{align}
\argmin_\theta \argmax_{\phi} \mathcal{L}_{\text{GAIL}}(\theta, \phi) = \mathbb{E}_{\pi_\theta} \left[\log D_\phi(s,a) \right] + \mathbb{E}_{\pi_E} \left[ \log (1 - D_\phi(s, a))\right] - \lambda \mathcal{H}(\pi_\theta)
\end{align}
where $\pi_\theta,\pi_E$ are a policy parameterized by $\theta$ and an expert policy respectively. $D_\phi(s,a):\mathcal{S}\times\mathcal{A} \rightarrow [0, 1]$ is a discriminator parameterized by $\phi$. $\mathcal{H}(\pi)=\mathbb{E}_{\pi}\left[-\log \pi(a|s)\right]$ is the entropy of the policy. 
Similar to GANs, the discriminator and the policy play an adversarial game by either maximizing or minimizing the objective $\mathcal{L}_{\text{GAIL}}$, and the gradient of each component is given by:
\begin{align}
\nabla_{\phi} \mathcal{L}_{\text{GAIL}} &= \mathbb{E}_{\tau_\pi} \left[\nabla_\phi \log D_\phi(s,a) \right] + \mathbb{E}_{\tau_E} \left[\nabla_\phi \log (1 - D_\phi(s, a))\right] 
\\
\nabla_\theta \mathcal{L}_{\text{GAIL}} &= \mathbb{E}_{\tau_\pi} \left[\nabla_\theta \log D_\phi(s,a) \right] - \lambda \mathcal{H}(\pi_\theta)
\label{eq:gail-policy-grad1}
\\
&=  \mathbb{E}_{\tau_\pi} \left[\nabla_\theta \log\pi_\theta(a|s) Q(s,a) \right] - \lambda \mathcal{H}(\pi_\theta),
\label{eq:gail-policy-grad2}
\end{align}
where $Q(s,a) = \mathbb{E}_{\tau_\pi}\left[\log D_\phi(s,a) | s_0=s,a_0=a \right]$, and $\tau_\pi,\tau_E$ are trajectories sampled from $\pi_\theta$ and $\pi_E$ respectively.
Intuitively, the discriminator $D_\phi$ is trained to discriminate between the policy's trajectories ($\tau_\pi$) and the expert's trajectories ($\tau_E$) through cross entropy loss. On the other hand, the policy $\pi_\theta$ is trained to fool the discriminator by generating trajectories that are close to the expert trajectories according to the discriminator. Since $\log D_\phi(s,a)$ is non-differentiable with respect to $\theta$ in Equation~\ref{eq:gail-policy-grad1}, the score function estimator is used to compute the gradient, which leads to a form of policy gradient (Equation~\ref{eq:gail-policy-grad2}) using the discriminator as a reward function.

It has been shown that GAIL amounts to minimizing the Jensen-Shannon divergence between the policy's occupancy measure and the expert's~\citep{ho2016generative,goodfellow2014generative} as follows:
\begin{align}
\argmin_\theta \argmax_{\phi} \mathcal{L}_{\text{GAIL}}(\theta, \phi)= \argmin_\theta D_\text{JS}(\rho_{\pi_\theta} || \rho_{\pi_E}) -\lambda \mathcal{H}(\pi_\theta)
\end{align}
where $D_\text{JS}(p || q)=D_\text{KL}(p || (p+q)/2)+D_\text{KL}(q || (p+q)/2)$ denotes Jensen-Shannon divergence, a distance metric between two distributions, which is minimized when $p=q$. 

\cutsectionup
\section{\gasil{}} \label{sec:method}
\cutsectiondown
The main idea of \gasil{} (\GASIL{}) is to update the policy to imitate past good trajectories using GAIL framework (see Section~\ref{sec:gail} for GAIL). We describe the details of \GASIL{} in Section~\ref{sec:gasil-algorithm} and make a connection between \GASIL{} and reward learning in Section~\ref{sec:gasil-reward-learning}, which leads to a combination of \GASIL{} with policy gradient in Section~\ref{sec:gasil-combination}.

\begin{algorithm}[t]
\caption{\gasil{}}\label{alg:gasil}
\begin{algorithmic}
\State Initialize policy parameter $\theta$
\State Initialize discriminator parameter $\phi$
\State Initialize good trajectory buffer $\mathcal{B} \gets \emptyset$
\For{each iteration}
    \State Sample policy trajectories $\tau_\pi \sim \pi_\theta$ 
    \State Update good trajectory buffer $\mathcal{B}$ using $\tau_\pi$ 
    \State Sample good trajectories $\tau_E \sim \mathcal{B}$ 
    \State Update the discriminator parameter $\phi$ via gradient ascent with:
    \begin{equation}
\nabla_\phi \mathcal{L}_{\text{\GASIL{}}} = \mathbb{E}_{\tau_\pi} \left[\nabla_\phi \log D_\phi(s,a) \right] + \mathbb{E}_{\tau_E} \left[\nabla_\phi \log (1 - D_\phi(s, a))\right]
    \end{equation}
	\State Update the policy parameter $\theta$ via gradient descent with:
	\begin{equation}
    \begin{aligned}
& \nabla_\theta \mathcal{L}_{\text{\GASIL{}}} = \mathbb{E}_{\tau_\pi} \left[\nabla_\theta \log\pi_\theta(a|s) Q(s,a) \right] - \lambda \nabla_\theta \mathcal{H}(\pi_\theta), \\
& \text{where } Q(s,a) = \mathbb{E}_{\tau_\pi}\left[\log D_\phi(s,a) | s_0=s,a_0=a \right] 
    \end{aligned}
    \label{eq:gasil-policy}
    \end{equation}
\EndFor
\end{algorithmic}
\end{algorithm}


\cutsubsectionup
\subsection{Algorithm} \label{sec:gasil-algorithm}
\cutsubsectiondown
The keay idea of \GASIL{} is to treat good trajectories collected by the agent as trajectories that the agent should imitate as described in Algorithm~\ref{alg:gasil}. More specifically, \GASIL{} performs the following two updates for each iteration.

\paragraph{Updating good trajectory buffer ($\mathcal{B}$)} 
\GASIL{} maintains a \textit{good trajectory buffer} $\mathcal{B}=\{\tau_i\}$ that contains a few trajectories ($\tau_i$) that obtained high rewards in the past. Each trajectory consists of a sequence of states and actions: $s_0,a_0,s_1,a_1,...,s_T$. We define `good trajectories' as any trajectories whose the discounted sum of rewards are higher than that of the policy. Though there can be many different ways to obtain such trajectories, we propose to store top-K episodes according to the return $R=\sum^\infty_{t=0}\gamma^t r_t$. 

\paragraph{Updating discriminator ($D_\phi$) and policy ($\pi_\theta$)} 
The agent learns to imitate good trajectories contained in the good trajectory buffer $\mathcal{B}$ using generative adversarial imitation learning. More formally, the discriminator ($D_\phi(s,a)$) and the policy ($\pi_\theta(a|s)$) are updated via the following objective:
\begin{align}
\argmin_\theta \argmax_{\phi} \mathcal{L}_{\text{\GASIL{}}}(\theta, \phi) = \mathbb{E}_{\tau_{\pi}} \left[\log D_\phi(s,a) \right] + \mathbb{E}_{\tau_E \sim \mathcal{B}} \left[ \log (1 - D_\phi(s, a))\right] - \lambda \mathcal{H}(\pi_\theta)
\label{eq:gasil-objective}
\end{align}
where $\tau_\pi,\tau_E$ are sampled trajectories from the policy $\pi_\theta$ and the good trajectory buffer $\mathcal{B}$ respectively. Intuitively, the discriminator is trained to discriminate between good trajectories and the policy's trajectories, while the policy is trained to make it difficult for the discriminator to distinguish by imitating good trajectories.

\cutsubsectionup
\subsection{Connection to Reward Learning}
\label{sec:gasil-reward-learning}
\cutsubsectiondown
The discriminator in \GASIL{} can be interpreted as a reward function for which the policy optimizes because Equation~\ref{eq:gasil-policy} uses the score function estimator to maximize the reward given by $-\log D_\phi(s,a)$. In other words, the policy is updated to maximize the discounted sum of rewards given by the discriminator rather than the true reward from the environment. Since the discriminator is also \textit{learned}, \GASIL{} can be viewed as an instance of optimal reward learning algorithm~\citep{sorg2010reward}. A potential benefit of \GASIL{} is that the optimal discriminator can provide intermediate rewards to the policy along good trajectories, even if the true reward from the environment is delayed or sparse. In such a scenario, \GASIL{} can allow the agent to learn more easily compared to the true reward function. Indeed, as we will show in Section~\ref{sec:delayed-mujoco}, \GASIL{} performs significantly better than a state-of-the-art policy gradient baseline in a delayed reward setting.

\cutsubsectionup
\subsection{Combining with Policy Gradient} \label{sec:gasil-combination}
\cutsubsectiondown
As the discriminator can interpreted as a learned internal reward function, it can be easily combined with any RL algorithms. In this paper, we explore a combination of \GASIL{} objective and policy gradient objective (Equation~\ref{eq:pg}) by performing a gradient ascent with the following accumulated gradient:
\begin{align}
\nabla_\theta J_{\text{PG}} - \alpha \nabla_\theta\mathcal{L}_{\text{\GASIL{}}} = \mathbb{E}_{\pi_\theta}\left[\nabla_\theta \log \pi_\theta(a|s)\hat{A}^{\alpha}_t \right]
\end{align}
where $\hat{A}^{\alpha}_t$ is an advantage estimation using a modified reward function $r^{\alpha}(s,a) \triangleq r(s,a) - \alpha \log D_\phi(s,a)$. Intuitively, the discriminator is used to shape the reward function to encourage the policy to imitate good trajectories. 

\cutsectionup
\section{Experiments}
\cutsectiondown
The experiments are designed to answer the following questions: (1) What is learned by \GASIL{}? (2) Is \GASIL{} better than behavior cloning approach? (3) Is \GASIL{} competitive to policy gradient method?; (4) Is \GASIL{} complementary to policy gradient method when combined together?

\cutsubsectionup
\subsection{Implementation Details}
\cutsubsectiondown
We implemented the following agents:
\begin{itemize}[leftmargin=4mm]
\item PPO: Proximal policy optimization (PPO) baseline~\citep{Schulman2017ProximalPO}.
\item PPO+BC: PPO with additional behavior cloning to top-K trajectories.
\item PPO+SIL: PPO with self-imitation learning~\citep{oh2018self}.
\item PPO+\GASIL{}: Our method using both the discriminator and the true reward (Section~\ref{sec:gasil-combination}).
\end{itemize}

The details of the network architectures and hyperparameters are described in the appendix. Our implementation is based on OpenAI's PPO and GAIL implementations~\citep{baselines}. 

\begin{wrapfigure}{r}{0.3\textwidth}
	\vskip -0.1in
	\small
    \centering
	\includegraphics[width=1.0\linewidth]{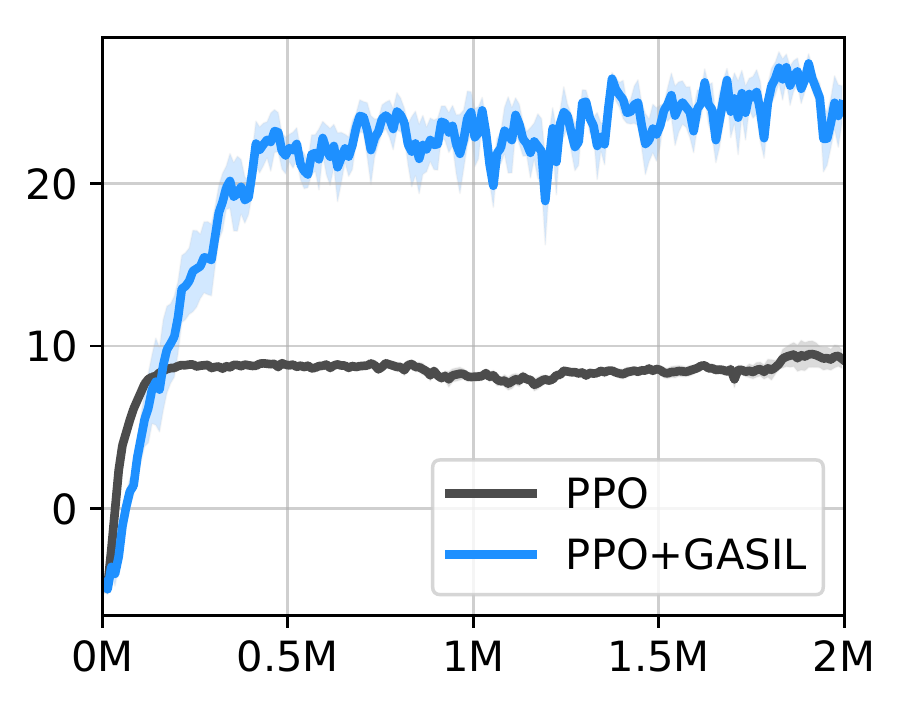}
	\caption{Learning curve on 2D point mass. See text for details. } 
	\label{fig:mass}	
	\vskip -0.1in
\end{wrapfigure}

\cutsubsectionup
\subsection{2D Point Mass} 
\cutsubsectiondown
To better understand how \GASIL{} works, we implemented a simple 2D point mass environment with continuous actions that determine the velocity of the agent in a 2D space as illustrated in Figure~\ref{fig:point-mass-gasil}. In this environment, the agent should collect as many blue/green objects as possible that give positive rewards (5 and 10 respectively) while avoiding distractor objects (orange) that give negative rewards (-5). There is also an actuation cost proportional to L2-norm of action which discourages large velocity.

The result in Figure~\ref{fig:mass} shows that PPO tends to learn a sub-optimal policy quickly. Although PPO+\GASIL{} tends to learn slightly slowly at the early stage, it finds a better policy at the end of learning compared to PPO. 

Figure~\ref{fig:point-mass-gasil} visualizes the learning progress of \GASIL{} with the learned discriminator. It is shown that the initial top-K trajectories collect several positive objects as well as distractors on the top area of the environment. This encourages the policy to explore the top area because \GASIL{} encourages the agent to imitate those top-K trajectories. As visualized in the third row in Figure~\ref{fig:point-mass-gasil}, the discriminator learns to put higher rewards for state-actions that are close to top-k trajectories, which strongly encourages the policy to imitate such trajectories. As training goes and the policy improves, the agent finds better trajectories that avoid distractors while collecting positive rewards. The good trajectory buffer is updated accordingly as the agent collects such trajectories, which is used to train the discriminator. The interaction between the policy and the discriminator converges to a sub-optimal policy which collects two green objects.

In contrast, Figure~\ref{fig:point-mass-ppo} visualizes the learning progress of PPO. Even though the agent collected the same top-k trajectories at the beginning as in PPO+\GASIL{} (compare the first columns of Figure~\ref{fig:point-mass-gasil} and Figure~\ref{fig:point-mass-ppo}), the policy trained with PPO objective quickly converges to a sub-optimal policy which collects only one green object depending on initial positions. We conjecture that this is because the policy gradient objective (Eq~\ref{eq:pg}) with the true reward function strongly encourages collecting nearby positive rewards and discourages collecting negative rewards. Thus, once the agent learns a sub-optimal behavior as shown in Figure~\ref{fig:point-mass-ppo}, the true reward function discourages further exploration due to distractors (orange objects) nearby green objects and the actuation cost. 
 
On the other hand, our \GASIL{} objective does not explicitly encourage nor discourage the agent to collect positive/negative rewards, because the discriminator gives the agent internal rewards according to how close the agent's trajectories are to top-K trajectories regardless of whether it collects some objects or not. Though this can possibly slow down learning, it can often help finding a better policy in the end depending on tasks as shown in this experiment. This result also implies that directly learning to maximize true reward such as policy gradient method may not be always optimal for learning a desired behavior.

\newcommand{\subwidth}{0.180}
\newcommand{\twidth}{2cm}
\begin{figure}[t]
     \small
     \setlength{\tabcolsep}{1pt}
     \centering
     \begin{tabular}{clllll}
     \rotatebox{90}{\hspace{-1.2cm} Policy trajectories}
     & 
     \begin{subfigure}{\subwidth\linewidth}
 	    \includegraphics[width=1\linewidth]{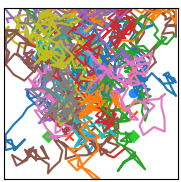} 
 	\end{subfigure} 
 	& 
     \begin{subfigure}{\subwidth\linewidth}
 	    \includegraphics[width=1\linewidth]{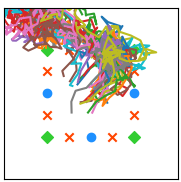} 
 	\end{subfigure} 
 	& 
 	\begin{subfigure}{\subwidth\linewidth}
 	    \includegraphics[width=1\linewidth]{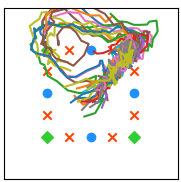} 
 	\end{subfigure} 
 	&
 	\begin{subfigure}{\subwidth\linewidth}
 	    \includegraphics[width=1\linewidth]{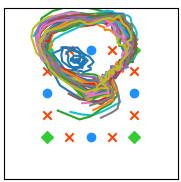} 
 	\end{subfigure} 
 	&
 	\begin{subfigure}{\subwidth\linewidth}
 	    \includegraphics[width=1\linewidth]{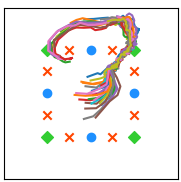} 
 	\end{subfigure} 
 	\\
 	\rotatebox{90}{\hspace{-1.2cm} Top-K trajectories}
     & 
 	\begin{subfigure}{\subwidth\linewidth}
 	    \includegraphics[width=1\linewidth]{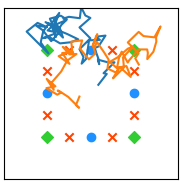} 
 	\end{subfigure} 
 	& 
     \begin{subfigure}{\subwidth\linewidth}
 	    \includegraphics[width=1\linewidth]{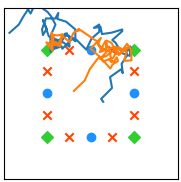} 
 	\end{subfigure} 
 	& 
 	\begin{subfigure}{\subwidth\linewidth}
 	    \includegraphics[width=1\linewidth]{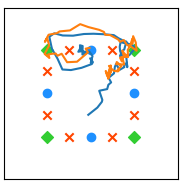} 
 	\end{subfigure} 
 	& 
 	\begin{subfigure}{\subwidth\linewidth}
 	    \includegraphics[width=1\linewidth]{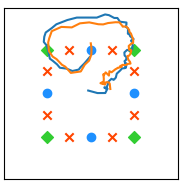} 
 	\end{subfigure} 
 	&
 	\begin{subfigure}{\subwidth\linewidth}
 	    \includegraphics[width=1\linewidth]{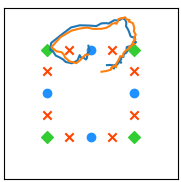} 
 	\end{subfigure} 
 	\\
 	\rotatebox{90}{\hspace{-1.0cm}\parbox{\twidth}{\centering Discriminator reward}}
     & 
 	\begin{subfigure}{\subwidth\linewidth}
 	    \includegraphics[width=1\linewidth]{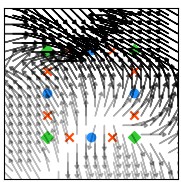} 
 	\end{subfigure} 
 	& 
     \begin{subfigure}{\subwidth\linewidth}
 	    \includegraphics[width=1\linewidth]{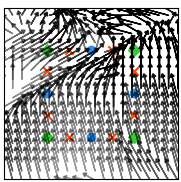} 
 	\end{subfigure} 
 	& 
 	\begin{subfigure}{\subwidth\linewidth}
 	    \includegraphics[width=1\linewidth]{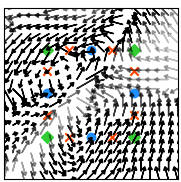} 
 	\end{subfigure} 
 	&
 	\begin{subfigure}{\subwidth\linewidth}
 	    \includegraphics[width=1\linewidth]{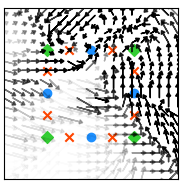} 
 	\end{subfigure} 
 	&
 	\begin{subfigure}{\subwidth\linewidth}
 	    \includegraphics[width=1\linewidth]{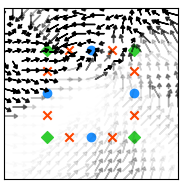} 
 	\end{subfigure} 
 	\\
 	\end{tabular}
 	 \vskip -0.05in
     \caption{Visualization of \GASIL{} policy on 2D point mass. The first two rows show the agent's trajectories and top-k trajectories at different training steps from left to right. The third row visualizes the learned discriminator at the corresponding training steps. Each arrow shows the best action at each position of the agent for which the discriminator gives the highest reward. The transparency of each arrow represents the magnitude of the discriminator reward (higher transparency represents lower reward).} 
 	\label{fig:point-mass-gasil}
 	\vskip -0.1in
\end{figure}
\begin{figure}[t]
     \small
     \setlength{\tabcolsep}{1pt}
     \centering
     \begin{tabular}{clllll}
     \rotatebox{90}{\hspace{-1.2cm} Policy trajectories}
     & 
 	\begin{subfigure}{\subwidth\linewidth}
 	    \includegraphics[width=1\linewidth]{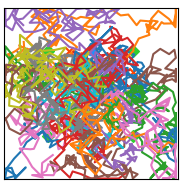} 
 	\end{subfigure} 
 	& 
     \begin{subfigure}{\subwidth\linewidth}
 	    \includegraphics[width=1\linewidth]{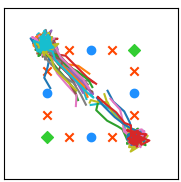} 
 	\end{subfigure} 
 	& 
 	\begin{subfigure}{\subwidth\linewidth}
 	    \includegraphics[width=1\linewidth]{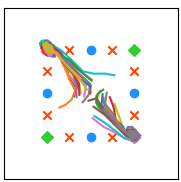} 
 	\end{subfigure} 
 	& 
 	\begin{subfigure}{\subwidth\linewidth}
 	    \includegraphics[width=1\linewidth]{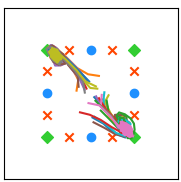} 
 	\end{subfigure} 
 	&
 	\begin{subfigure}{\subwidth\linewidth}
 	    \includegraphics[width=1\linewidth]{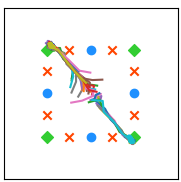} 
 	\end{subfigure} 
 	\\
 	\rotatebox{90}{\hspace{-1.2cm} Top-K trajectories}
     & 
 	\begin{subfigure}{\subwidth\linewidth}
 	    \includegraphics[width=1\linewidth]{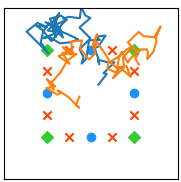} 
 	\end{subfigure} 
 	& 
     \begin{subfigure}{\subwidth\linewidth}
 	    \includegraphics[width=1\linewidth]{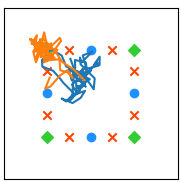} 
 	\end{subfigure} 
 	& 
 	\begin{subfigure}{\subwidth\linewidth}
 	    \includegraphics[width=1\linewidth]{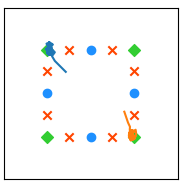} 
 	\end{subfigure} 
 	& 
 	\begin{subfigure}{\subwidth\linewidth}
 	    \includegraphics[width=1\linewidth]{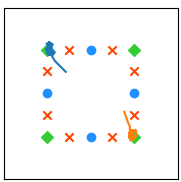} 
 	\end{subfigure} 
 	&
 	\begin{subfigure}{\subwidth\linewidth}
 	    \includegraphics[width=1\linewidth]{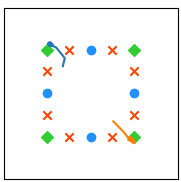} 
 	\end{subfigure} 
 	\\
 	\end{tabular}
 	\vskip -0.05in
     \caption{Visualization of PPO policy on 2D point mass. Compared to GASIL (Figure~\ref{fig:point-mass-gasil}), PPO tends to prematurely learn a worse policy. 
     } 
 	\label{fig:point-mass-ppo}
 	\vskip -0.15in
\end{figure}

\cutsubsectionup
\subsection{MuJoCo} \label{sec:mujoco}
\begin{figure}
    \small
    \centering
    \includegraphics[width=0.9\linewidth]{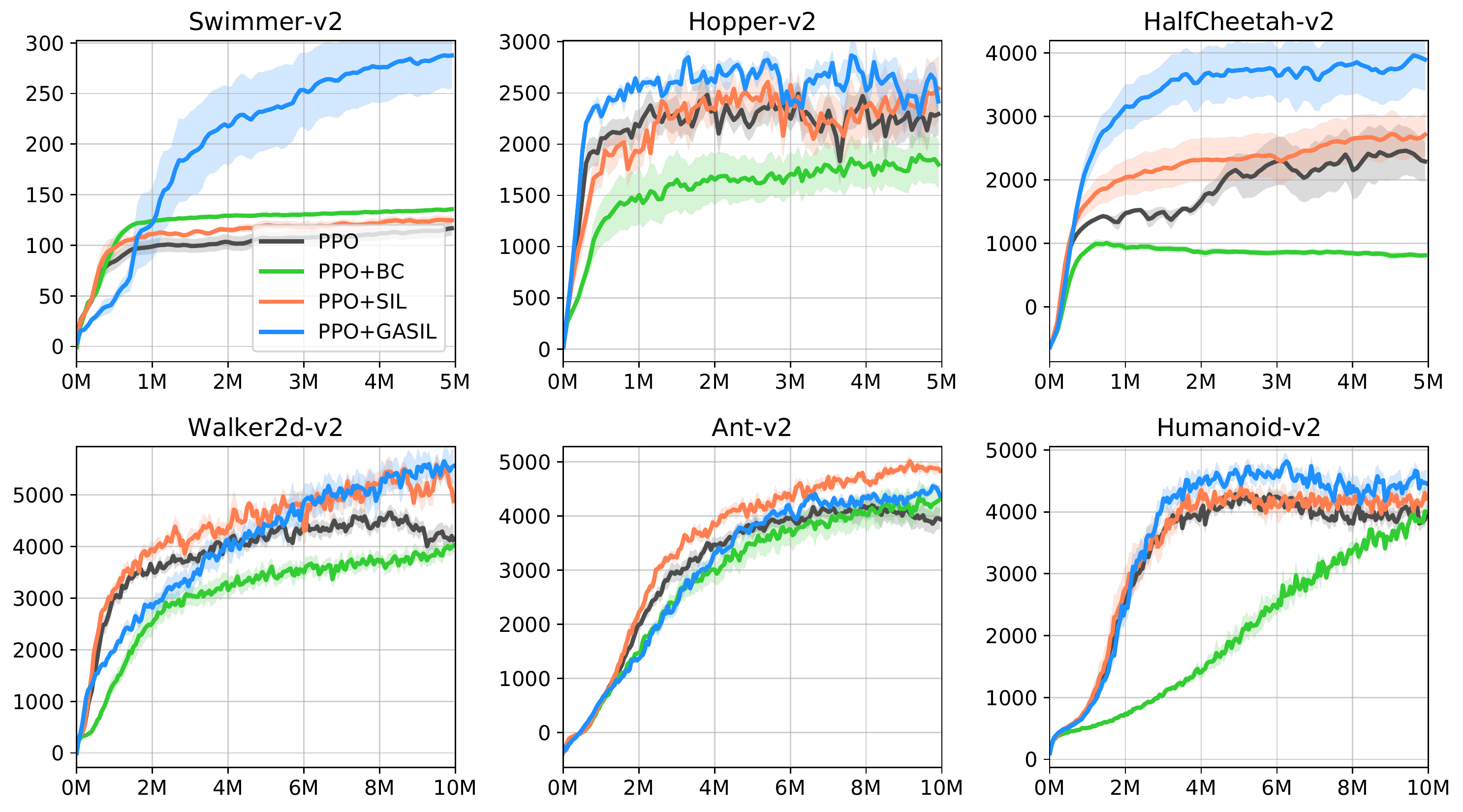} 
    \vskip -0.05in
    \caption{Learning curves on OpenAI Gym MuJoCo tasks averaged over 10 independent runs. x-axis and y-axis correspond to the number of steps and average reward. }
    \label{fig:mujoco}
    \vskip -0.1in
\end{figure}

\cutsubsectiondown
To further investigate how well \GASIL{} performs on complex control tasks, we evaluated it on OpenAI Gym MuJoCo tasks~\citep{brockman2016openai,Todorov2012MuJoCoAP}.\footnote{The demo video of the learned policies are available at \url{https://youtu.be/zvSr9gYEgGo}.} 
The result in Figure~\ref{fig:mujoco} shows that \GASIL{} improves PPO on most of the tasks. This indicates that \GASIL{} objective can be complementary to PPO objective, and the learned reward acts as a useful reward shaping that makes learning easier.

It is also shown that \GASIL{} significantly outperforms the behavior cloning baseline (`PPO+BC') on most of the tasks. Behavior cloning has been shown to require a large amount of samples to imitate compared to GAIL as shown by~\citet{ho2016generative}. This can be even more crucial in the RL context because there are not many good trajectories in the buffer (e.g., 1K-10K samples). Besides, \GASIL{} also outperforms self-imitation learning (`PPO+SIL')~\citep{oh2018self} showing that our generative adversarial approach is more sample-efficient than self-imitation learning. In fact, self-imitation learning can be viewed as an advantaged-weighted behavior cloning with prioritized replay, which can be the reason why \GASIL{} is more sample-efficient. Another possible reason would be that \GASIL{} generalizes better than behavior cloning method under the non-stationary data (i.e., good trajectory buffer). 

\begin{figure}
    \small
    \centering
    \includegraphics[width=0.98\linewidth]{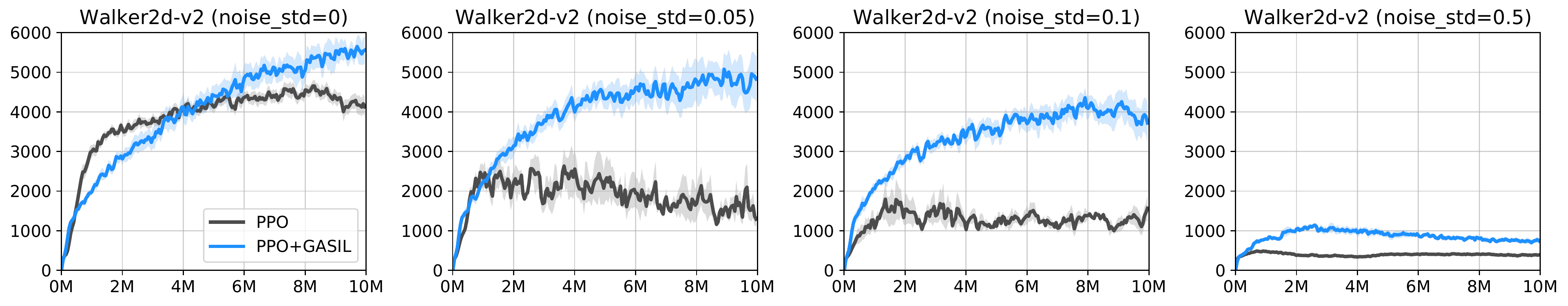} 
    \vskip -0.05in
    \caption{Learning curves on stochastic Walker2d-v2 averaged over 10 independent runs. The leftmost plot shows the learning curves on the original task without any noise in the environment. The other plots show learning curves on stochastic Walker2d-v2 task where Gaussian noise with standard deviation of $\{0.05, 0.1, 0.5\}$ (from left to right) is added to the observation for each step independently. }
    \label{fig:mujoco-stochastic}
    \vskip -0.1in
\end{figure}

We further investigated how robust \GASIL{} is to the stochasticity of the environment by adding a Gaussian noise to the observation for each step on Walker2d-v2. The result in Figure~\ref{fig:mujoco-stochastic} shows that the gap between PPO and PPO+GASIL is larger when the noise is added to the environment. This result suggests that \GASIL{} can be robust to stochastic environments to some extent in practice. 

\cutsubsectionup
\subsection{Delayed MuJoCo} \label{sec:delayed-mujoco}
\cutsubsectiondown
OpenAI Gym MuJoCo tasks provide dense reward signals to the agent according to the agent's progress along desired directions. In order to see how useful \GASIL{} is under more challenging reward structures, we modified the tasks by delaying the reward of MuJoCo tasks for 20 steps. In other words, the agent receives an accumulated reward only after every 20 steps or when the episode terminates. This modification makes it much more difficult for the agent to learn due to the delayed reward signal.

\begin{figure}
    \small
    \centering
    \includegraphics[width=0.9\linewidth]{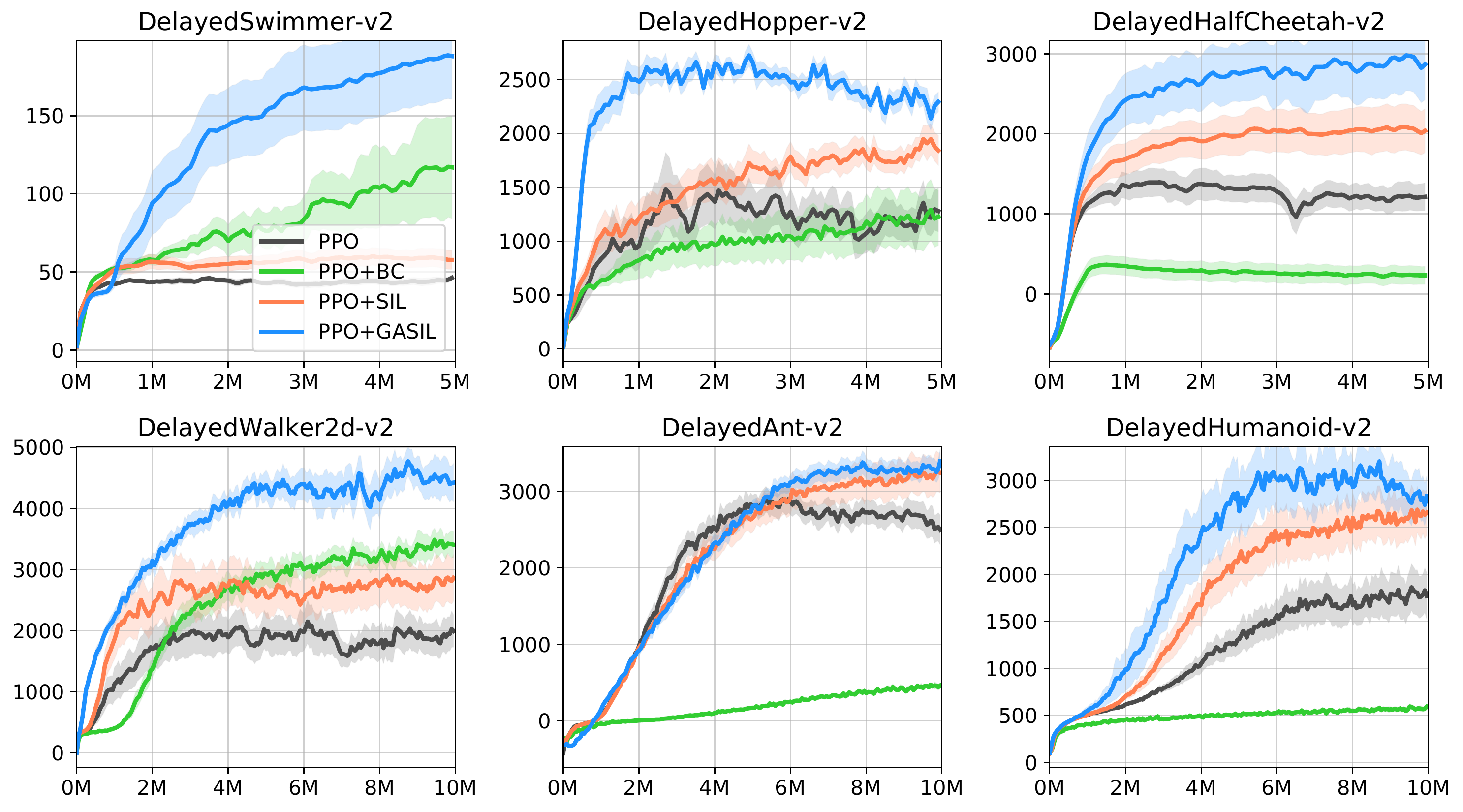} 
    \vskip -0.05in
    \caption{Learning curves on delayed-reward versions of OpenAI Gym MuJoCo tasks averaged over 10 independent runs. x-axis and y-axis correspond to the number of steps and average reward. }
    \label{fig:mujoco-delayed}
\end{figure}

The result in Figure~\ref{fig:mujoco-delayed} shows that \GASIL{} is much more helpful on delayed-reward MuJoCo tasks compared to non-delayed ones in Figure~\ref{fig:mujoco}, improving PPO on all tasks by a large margin. This result demonstrates that \GASIL{} is useful especially for dealing with delayed reward because the discriminator gives dense reward signals to the policy, even though the true reward is extremely delayed.

\cutsubsectionup
\subsection{Effect of hyperparameters}
\begin{figure}
    \small
    \centering
    \begin{subfigure}{0.3\linewidth}
 	    \includegraphics[width=1\linewidth]{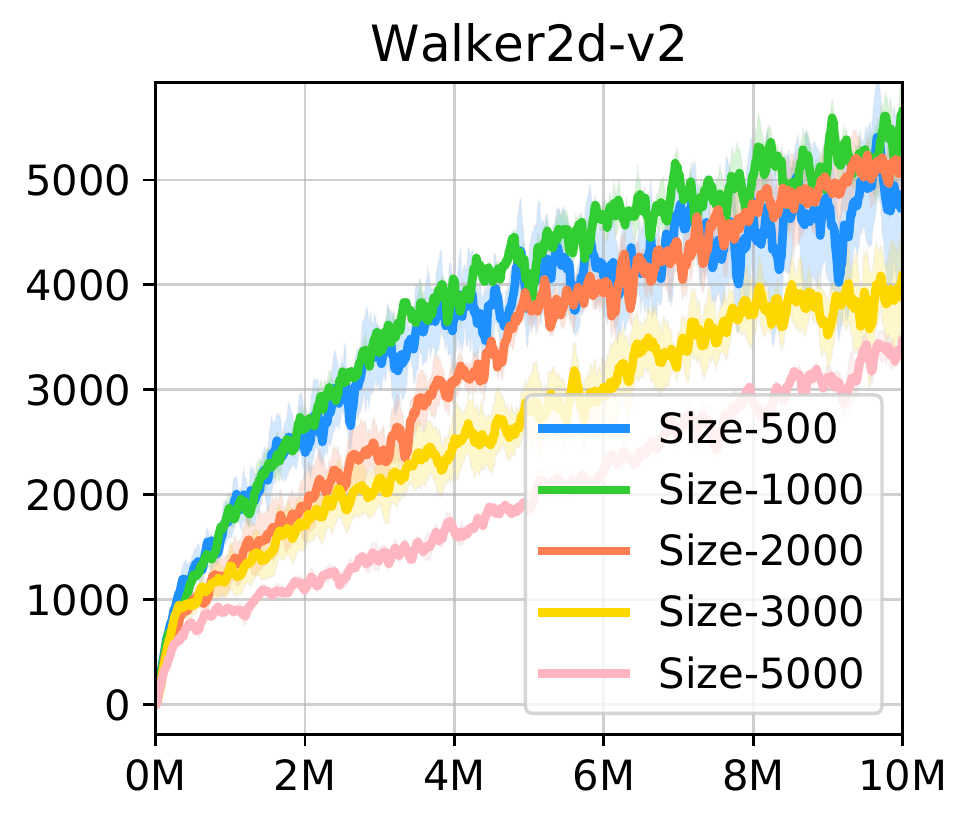} 
 	    \caption{Buffer size}
 	    \label{fig:hyper-buffer}
 	\end{subfigure} 
 	    \begin{subfigure}{0.3\linewidth}
 	    \includegraphics[width=1\linewidth]{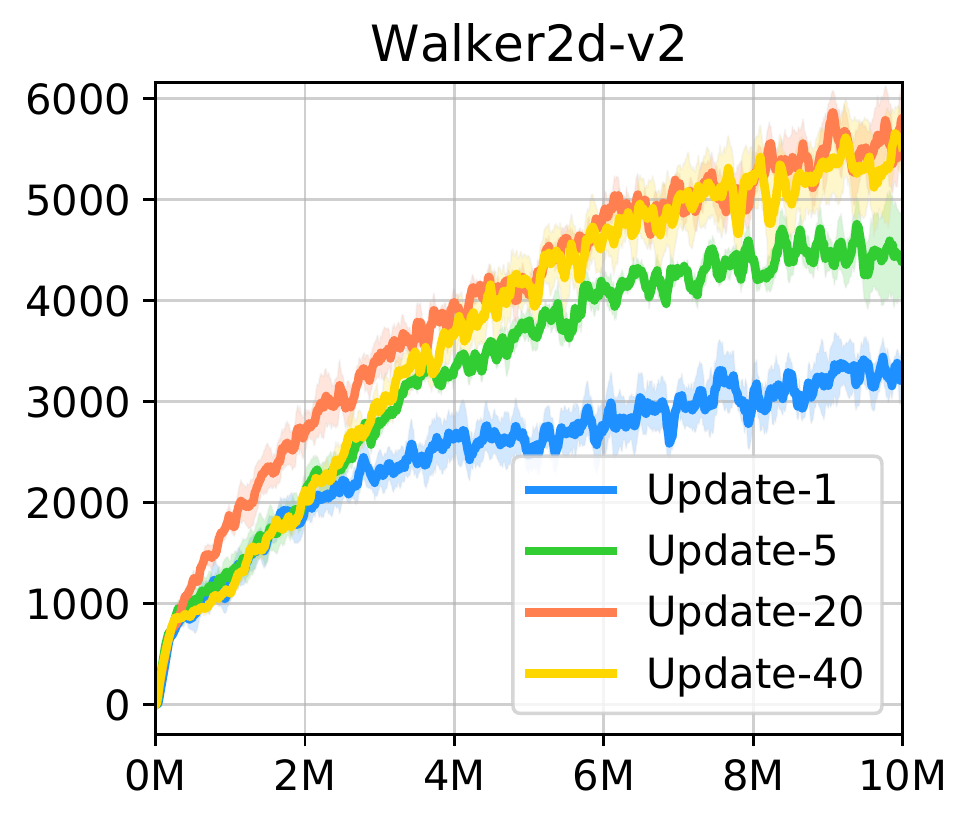} 
 	    \caption{Discriminator updates}
 	    \label{fig:discriminator-update}
 	\end{subfigure} 
 	\vskip -0.05in
 	\caption{Effect of \GASIL{} hyperparameters. }
 	\label{fig:hyper}
 	\vskip -0.1in
 \end{figure}

\cutsubsectiondown
Figure~\ref{fig:hyper} shows the effect of \GASIL{} hyperparameters on Walker2d-v2. Specifically, Figure~\ref{fig:hyper-buffer} shows the effect of the size of good trajectory buffer in terms of maximum steps in the buffer. It turns out that the agent performs poorly when the buffer size is too small (500 steps) or large (5000 steps). Although it is always useful to have more samples for imitation learning in general, the average return of good trajectories decreases as the size of the buffer increases. This indicates that there is a trade-off between the number of samples and the quality of good trajectories.

Figure~\ref{fig:discriminator-update} shows the effect of the number of discriminator updates with a fixed number of policy updates per batch. It is shown that too small or too large number of discriminator updates hurts the performance. This result is also consistent with GANs~\citep{goodfellow2014generative}, where the balance between the discriminator and the generator (i.e., policy) is crucial for the performance.

\cutsectionup
\section{Discussions} \label{sec:discussion}
\cutsectiondown
\cutparagraphup
\paragraph{Alternative ways of training the discriminator}
We presented a simple way of training the discriminator: discriminating between top-K trajectories and policy trajectories. 
However, there can be many different ways of defining good trajectories and training the discriminator. 
Developing a more principled way of training the discriminator with strong theoretical guarantee would be an important future work.

\cutparagraphup
\paragraph{Dealing with multi-modal trajectories} In the experiment, we used a Gaussian policy with an independent covariance. This type of policy has been shown to have difficulties in learning diverse behaviors~\citep{haarnoja2018soft,Haarnoja2017ReinforcementLW}. In \GASIL{}, we observed that the good trajectory buffer ($\mathcal{B}$) often contain multi-modal trajectories because they are collected by different policies with different parameters over time. This makes it difficult for a Gaussian policy to imitate reliably. In fact, there has been recent methods~\citep{hausman2017multi,li2017infogail} designed to imitate multi-modal behaviors in the GAIL framework. We believe that combining such methods would further improve the performance. 
\cutparagraphup
\paragraph{Model-based approach} We used a model-free GAIL framework which requires policy gradient for training the policy. However, our idea can be extended to model-based GAIL (MGAIL)~\citep{baram2017end} where the policy is updated by directly backpropagating through a learned discriminator and a learned dynamics model. Since MGAIL has been shown to be more sample-efficient, we expect that a model-based counterpart of \GASIL{} would also improve the performance.

\cutsectionup
\section{Conclusion}
\cutsectiondown
This paper proposed \gasil{} (\GASIL{}) as a simple regularizer for RL. The main idea is to imitate good trajectories that the agent has collected using generative adversarial learning framework. We demonstrated that \GASIL{} significantly improves existing state-of-the-art baselines across many control tasks especially when rewards are delayed.
Extending this work towards a more principled generative adversarial learning approach with stronger theoretical guarantee would be an interesting research direction.

\bibliography{references}
\bibliographystyle{plainnat}
\clearpage
\input{appendix.tex}
\end{document}

%% file: appendix.tex
\appendix

\clearpage
\section{Hyperparameters}
Hyperperameters and architectures used for MuJoCo experiments are described in Table~\ref{tab:hyperparameter}. We performed a random search over the range of hyperparameters specified in Table~\ref{tab:hyperparameter}. For GASIL+PPO on Humanoid-v2, the policy is trained with PPO ($\alpha=0$) for the first 2M steps, and $\alpha$ is increased to 0.02 until 3M steps. For the rest of tasks including all delayed-MuJoCo tasks, we used used a fixed $\alpha$ throughout training.
\begin{table}[H]
\small
\centering
\caption{GARL hyperparameters on MuJoCo.}
\label{tab:hyperparameter}
\begin{tabular}{l l}
\toprule
Hyperparameters & Value \\
\midrule
Architecture & FC(64)-FC(64) \\
Learning rate & \{0.0003, 0.0001, 0.00005, 0.00003\} \\
Horizon & 2048 \\
Number of epochs & 10 \\
Minibatch size & 64 \\
Discount factor ($\gamma$) & 0.99 \\
GAE parameter & 0.95 \\
Entropy regularization coefficient ($\lambda$) & 0 \\
\midrule
Discriminator minibatch size & 128 \\
Number of discriminator updates per batch & \{1, 5, 10, 20\} \\
Discriminator learning rate & \{0.0003, 0.0001, 0.00002, 0.00001\} \\
Size of good trajectory buffer (steps) & \{1000, 10000\} \\
Scale of discriminator reward ($\alpha$) & \{0.02, 0.1, 0.2, 1\} \\
\bottomrule
\end{tabular}
\end{table} 

%% file: main.bbl
\begin{thebibliography}{31}
\providecommand{\natexlab}[1]{#1}
\providecommand{\url}[1]{\texttt{#1}}
\expandafter\ifx\csname urlstyle\endcsname\relax
  \providecommand{\doi}[1]{doi: #1}\else
  \providecommand{\doi}{doi: \begingroup \urlstyle{rm}\Url}\fi

\bibitem[Abolafia et~al.(2018)Abolafia, Norouzi, and Le]{abolafia2018neural}
Daniel~A Abolafia, Mohammad Norouzi, and Quoc~V Le.
\newblock Neural program synthesis with priority queue training.
\newblock \emph{arXiv preprint arXiv:1801.03526}, 2018.

\bibitem[Baram et~al.(2017)Baram, Anschel, Caspi, and Mannor]{baram2017end}
Nir Baram, Oron Anschel, Itai Caspi, and Shie Mannor.
\newblock End-to-end differentiable adversarial imitation learning.
\newblock In \emph{International Conference on Machine Learning}, pages
  390--399, 2017.

\bibitem[Blundell et~al.(2016)Blundell, Uria, Pritzel, Li, Ruderman, Leibo,
  Rae, Wierstra, and Hassabis]{blundell2016model}
Charles Blundell, Benigno Uria, Alexander Pritzel, Yazhe Li, Avraham Ruderman,
  Joel~Z Leibo, Jack Rae, Daan Wierstra, and Demis Hassabis.
\newblock Model-free episodic control.
\newblock \emph{arXiv preprint arXiv:1606.04460}, 2016.

\bibitem[Brockman et~al.(2016)Brockman, Cheung, Pettersson, Schneider,
  Schulman, Tang, and Zaremba]{brockman2016openai}
Greg Brockman, Vicki Cheung, Ludwig Pettersson, Jonas Schneider, John Schulman,
  Jie Tang, and Wojciech Zaremba.
\newblock Openai gym.
\newblock \emph{arXiv preprint arXiv:1606.01540}, 2016.

\bibitem[Dhariwal et~al.(2017)Dhariwal, Hesse, Klimov, Nichol, Plappert,
  Radford, Schulman, Sidor, and Wu]{baselines}
Prafulla Dhariwal, Christopher Hesse, Oleg Klimov, Alex Nichol, Matthias
  Plappert, Alec Radford, John Schulman, Szymon Sidor, and Yuhuai Wu.
\newblock Openai baselines.
\newblock \url{https://github.com/openai/baselines}, 2017.

\bibitem[Gangwani et~al.(2018)Gangwani, Liu, and Peng]{gangwani2018learning}
Tanmay Gangwani, Qiang Liu, and Jian Peng.
\newblock Learning self-imitating diverse policies.
\newblock \emph{arXiv preprint arXiv:1805.10309}, 2018.

\bibitem[Goodfellow et~al.(2014)Goodfellow, Pouget-Abadie, Mirza, Xu,
  Warde-Farley, Ozair, Courville, and Bengio]{goodfellow2014generative}
Ian Goodfellow, Jean Pouget-Abadie, Mehdi Mirza, Bing Xu, David Warde-Farley,
  Sherjil Ozair, Aaron Courville, and Yoshua Bengio.
\newblock Generative adversarial nets.
\newblock In \emph{Advances in neural information processing systems}, pages
  2672--2680, 2014.

\bibitem[Goyal et~al.(2018)Goyal, Brakel, Fedus, Lillicrap, Levine, Larochelle,
  and Bengio]{goyal2018recall}
Anirudh Goyal, Philemon Brakel, William Fedus, Timothy Lillicrap, Sergey
  Levine, Hugo Larochelle, and Yoshua Bengio.
\newblock Recall traces: Backtracking models for efficient reinforcement
  learning.
\newblock \emph{arXiv preprint arXiv:1804.00379}, 2018.

\bibitem[Guo et~al.(2016)Guo, Singh, Lewis, and Lee]{Guo2016DeepLF}
Xiaoxiao Guo, Satinder~P. Singh, Richard~L. Lewis, and Honglak Lee.
\newblock Deep learning for reward design to improve monte carlo tree search in
  atari games.
\newblock In \emph{IJCAI}, 2016.

\bibitem[Haarnoja et~al.(2017)Haarnoja, Tang, Abbeel, and
  Levine]{Haarnoja2017ReinforcementLW}
Tuomas Haarnoja, Haoran Tang, Pieter Abbeel, and Sergey Levine.
\newblock Reinforcement learning with deep energy-based policies.
\newblock In \emph{ICML}, 2017.

\bibitem[Haarnoja et~al.(2018)Haarnoja, Zhou, Abbeel, and
  Levine]{haarnoja2018soft}
Tuomas Haarnoja, Aurick Zhou, Pieter Abbeel, and Sergey Levine.
\newblock Soft actor-critic: Off-policy maximum entropy deep reinforcement
  learning with a stochastic actor.
\newblock \emph{arXiv preprint arXiv:1801.01290}, 2018.

\bibitem[Hausman et~al.(2017)Hausman, Chebotar, Schaal, Sukhatme, and
  Lim]{hausman2017multi}
Karol Hausman, Yevgen Chebotar, Stefan Schaal, Gaurav Sukhatme, and Joseph~J
  Lim.
\newblock Multi-modal imitation learning from unstructured demonstrations using
  generative adversarial nets.
\newblock In \emph{Advances in Neural Information Processing Systems}, pages
  1235--1245, 2017.

\bibitem[Ho and Ermon(2016)]{ho2016generative}
Jonathan Ho and Stefano Ermon.
\newblock Generative adversarial imitation learning.
\newblock In \emph{Advances in Neural Information Processing Systems}, pages
  4565--4573, 2016.

\bibitem[Lengyel and Dayan(2008)]{lengyel2008hippocampal}
M{\'a}t{\'e} Lengyel and Peter Dayan.
\newblock Hippocampal contributions to control: the third way.
\newblock In \emph{Advances in neural information processing systems}, pages
  889--896, 2008.

\bibitem[Li et~al.(2017)Li, Song, and Ermon]{li2017infogail}
Yunzhu Li, Jiaming Song, and Stefano Ermon.
\newblock Infogail: Interpretable imitation learning from visual
  demonstrations.
\newblock In \emph{Advances in Neural Information Processing Systems}, pages
  3815--3825, 2017.

\bibitem[Liang et~al.(2016)Liang, Berant, Le, Forbus, and Lao]{liang2016neural}
Chen Liang, Jonathan Berant, Quoc Le, Kenneth~D Forbus, and Ni~Lao.
\newblock Neural symbolic machines: Learning semantic parsers on freebase with
  weak supervision.
\newblock \emph{arXiv preprint arXiv:1611.00020}, 2016.

\bibitem[Mansimov and Cho(2017)]{mansimov2017simple}
Elman Mansimov and Kyunghyun Cho.
\newblock Simple nearest neighbor policy method for continuous control tasks.
\newblock In \emph{NIPS Deep Reinforcement Learning Symposium}, 2017.

\bibitem[Ng et~al.(1999)Ng, Harada, and Russell]{ng1999policy}
Andrew~Y Ng, Daishi Harada, and Stuart Russell.
\newblock Policy invariance under reward transformations: Theory and
  application to reward shaping.
\newblock In \emph{ICML}, volume~99, pages 278--287, 1999.

\bibitem[Oh et~al.(2018)Oh, Guo, Singh, and Lee]{oh2018self}
Junhyuk Oh, Yijie Guo, Satinder Singh, and Honglak Lee.
\newblock Self-imitation learning.
\newblock In \emph{ICML}, 2018.

\bibitem[Pritzel et~al.(2017)Pritzel, Uria, Srinivasan, Puigdom{\`e}nech,
  Vinyals, Hassabis, Wierstra, and Blundell]{pritzel2017neural}
Alexander Pritzel, Benigno Uria, Sriram Srinivasan, Adri{\`a} Puigdom{\`e}nech,
  Oriol Vinyals, Demis Hassabis, Daan Wierstra, and Charles Blundell.
\newblock Neural episodic control.
\newblock \emph{arXiv preprint arXiv:1703.01988}, 2017.

\bibitem[Radford et~al.(2015)Radford, Metz, and
  Chintala]{radford2015unsupervised}
Alec Radford, Luke Metz, and Soumith Chintala.
\newblock Unsupervised representation learning with deep convolutional
  generative adversarial networks.
\newblock \emph{arXiv preprint arXiv:1511.06434}, 2015.

\bibitem[Reed et~al.(2016)Reed, Akata, Yan, Logeswaran, Schiele, and
  Lee]{reed2016generative}
Scott Reed, Zeynep Akata, Xinchen Yan, Lajanugen Logeswaran, Bernt Schiele, and
  Honglak Lee.
\newblock Generative adversarial text to image synthesis.
\newblock \emph{arXiv preprint arXiv:1605.05396}, 2016.

\bibitem[Schulman et~al.(2017)Schulman, Wolski, Dhariwal, Radford, and
  Klimov]{Schulman2017ProximalPO}
John Schulman, Filip Wolski, Prafulla Dhariwal, Alec Radford, and Oleg Klimov.
\newblock Proximal policy optimization algorithms.
\newblock \emph{CoRR}, abs/1707.06347, 2017.

\bibitem[Singh et~al.(2009)Singh, Lewis, and Barto]{singh2009rewards}
Satinder Singh, Richard~L Lewis, and Andrew~G Barto.
\newblock Where do rewards come from.
\newblock In \emph{Proceedings of the annual conference of the cognitive
  science society}, pages 2601--2606, 2009.

\bibitem[Sorg et~al.(2010)Sorg, Lewis, and Singh]{sorg2010reward}
Jonathan Sorg, Richard~L Lewis, and Satinder~P Singh.
\newblock Reward design via online gradient ascent.
\newblock In \emph{Advances in Neural Information Processing Systems}, pages
  2190--2198, 2010.

\bibitem[Syed et~al.(2008)Syed, Bowling, and Schapire]{syed2008apprenticeship}
Umar Syed, Michael Bowling, and Robert~E Schapire.
\newblock Apprenticeship learning using linear programming.
\newblock In \emph{Proceedings of the 25th international conference on Machine
  learning}, pages 1032--1039. ACM, 2008.

\bibitem[Todorov et~al.(2012)Todorov, Erez, and Tassa]{Todorov2012MuJoCoAP}
Emanuel Todorov, Tom Erez, and Yuval Tassa.
\newblock Mujoco: A physics engine for model-based control.
\newblock \emph{2012 IEEE/RSJ International Conference on Intelligent Robots
  and Systems}, pages 5026--5033, 2012.

\bibitem[Watkins and Dayan(1992)]{watkins1992q}
Christopher~JCH Watkins and Peter Dayan.
\newblock Q-learning.
\newblock \emph{Machine learning}, 8\penalty0 (3-4):\penalty0 279--292, 1992.

\bibitem[Zheng et~al.(2018)Zheng, Oh, and Singh]{zheng2018learning}
Zeyu Zheng, Junhyuk Oh, and Satinder Singh.
\newblock On learning intrinsic rewards for policy gradient methods.
\newblock \emph{arXiv preprint arXiv:1804.06459}, 2018.

\bibitem[Zhu et~al.(2017)Zhu, Park, Isola, and Efros]{zhu2017unpaired}
Jun-Yan Zhu, Taesung Park, Phillip Isola, and Alexei~A Efros.
\newblock Unpaired image-to-image translation using cycle-consistent
  adversarial networks.
\newblock \emph{arXiv preprint arXiv:1703.10593}, 2017.

\bibitem[Ziebart et~al.(2008)Ziebart, Maas, Bagnell, and
  Dey]{ziebart2008maximum}
Brian~D Ziebart, Andrew~L Maas, J~Andrew Bagnell, and Anind~K Dey.
\newblock Maximum entropy inverse reinforcement learning.
\newblock In \emph{AAAI}, volume~8, pages 1433--1438. Chicago, IL, USA, 2008.

\end{thebibliography}
